\documentclass{article}
\usepackage{silence}
\usepackage{etoolbox}

\usepackage{arxiv}
\usepackage{subcaption} 
\usepackage{natbib}

\usepackage[utf8]{inputenc}
\usepackage[T1]{fontenc}
\usepackage{microtype}  
\usepackage{CJK}

\usepackage{hyperref}       
\usepackage{url}            
\usepackage{booktabs}       
\usepackage{nicefrac}       
\usepackage{microtype}      
\usepackage{lipsum}
\usepackage{amsfonts,amssymb,amsbsy,textcomp,marvosym,amsmath,threeparttable,amsthm,float,lastpage,lscape}
\usepackage{eurosym,mathrsfs,fancyhdr,CJK,multicol,graphics,indentfirst,color,bm,upgreek,booktabs,graphicx,multirow}
\usepackage{warpcol}

\graphicspath{ {./images/} }

\title{MTOS: A LLM-Driven Multi-topic Opinion Simulation Framework for Exploring Echo Chamber Dynamics}

\author{
Dingyi Zuo$^{1}$,
Hongjie Zhang$^{1,*}$,
Jie Ou$^{2}$,
Chaosheng Feng$^{1}$,
Shuwan Liu$^{1}$\\[6pt]
$^{1}$Sichuan Normal University, College of Computer Science, Chengdu, China\\
$^{2}$University of Electronic Science and Technology of China,\\
School of Information and Software Engineering, Chengdu, China\\
$^{*}$\texttt{zhanghongjie@sicnu.edu.cn}
}

\begin{document}

\maketitle
\begin{abstract}
The polarization of opinions, information segregation, and cognitive biases on social media have attracted significant academic attention. In real-world networks, information often spans multiple interrelated topics, posing challenges for opinion evolution and highlighting the need for frameworks that simulate interactions among topics. Existing studies based on large language models (LLMs) focus largely on single topics, limiting the capture of cognitive transfer in multi-topic, cross-domain contexts. Traditional numerical models, meanwhile, simplify complex linguistic attitudes into discrete values, lacking interpretability, behavioral consistency, and the ability to integrate multiple topics. To address these issues, we propose Multi-topic Opinion Simulation (MTOS), a social simulation framework integrating multi-topic contexts with LLMs. MTOS leverages LLMs alongside short-term and long-term memory, incorporates multiple user-selection interaction mechanisms and dynamic topic-selection strategies, and employs a belief decay mechanism to enable perspective updates across topics. We conduct extensive experiments on MTOS, varying topic numbers, correlation types, and performing ablation studies to assess features such as group polarization and local consistency. Results show that multi-topic settings significantly alter polarization trends: positively correlated topics amplify echo chambers, negatively correlated topics inhibit them, and irrelevant topics also mitigate echo chamber effects through resource competition. Compared with numerical models, LLM-based agents realistically simulate dynamic opinion changes, reproduce linguistic features of news texts, and capture complex human reasoning, improving simulation interpretability and system stability.
\end{abstract}



\section{Introduction}
Polarization of views due to the echo chamber effect has become the core problem of group cognitive alienation in the social media environment. Individuals tend to be exposed to sources of information that agree with their own views and reject dissent, a phenomenon of ``information cocooning'' that is exacerbated by algorithmic recommendation mechanisms, leading to convergence of views, self-reinforcement, and social fragmentation \cite{Gromping2014, LevyRazin2019, TerrenBorgeBravo2021}. Modeling and intervening in this process has become an important topic in computational social science and artificial intelligence research. To simulate the process of opinion polarization, researchers have proposed various numerical models, such as the Friedkin--Johnsen model and the Bounded Confidence model \cite{FriedkinJohnsen1990, Deffuant2000}. Some extended versions also attempt to introduce multidimensional opinion vectors to simulate multi-topic contexts \cite{Lorenz2007, HegselmannKrause2014}. Although these models have advantages in interpretability and controllability, their modeling foundation still relies on compressing complex linguistic interactions and cognitive processes into numerical variables, making it difficult to capture the multi-topic interplay, cognitive transfer, and cross-topic associative abilities present in real social contexts.

Recently, with the rise of LLMs, a new class of frameworks that combine language generation with social simulation has emerged. For example, the Social Simulation Framework \cite{WangLiuYangChen2025} simulates natural language interactions among multiple agents by incorporating cognitive agents with language understanding and generation capabilities, and implements a text-based opinion updating mechanism. This model shows stronger anthropomorphic and explanatory power in portraying the evolution of opinions at the linguistic level, providing new ideas for the simulation of polarization mechanisms. S3 (Social Network Simulation System) \cite{gao2023s3} uses LLM-driven agents to simulate the emotions, attitudes, and interactive behaviors of individuals in social networks. SoMoSiMu (Social Media User Simulation) \cite{mou2024hisim} proposes a hybrid framework that categorizes social media users into core LLM-driven agents and ordinary deductive agents, and simulates user response dynamics after event triggering by constructing a Twitter-like environment. The method is evaluated by a multidimensional benchmark, SoMoSiMu-Bench, and its validity and flexibility are verified on a real dataset, providing a new approach to the simulation of social movement participant behavior. For example, some studies have proposed quantitative indicators to portray the competitive characteristics among topics \cite{SunZhouGuan2016}, while some scholars have established a multi-topic selection communication model and found that topic popularity has a significant impact on user selection behavior \cite{ChenZhangHuang2017}; further work has combined the relationship between users and topics to propose a heat prediction model and an attention shift prediction method \cite{yu2017, an2024}. These studies suggest that multi-topic crossover is an important factor influencing opinion dynamics, but there is still a lack of simulation frameworks that systematically portray multi-topic cognitive evolution at the level of language interaction.

However, existing LLM opinion simulations still suffer from a key limitation: all agents interact around a single topic only, ignoring the phenomena of multi-topic crossover, cognitive transfer, and cross-topic association in real social scenarios. Indeed, social discussions often span multiple domains, such as ethics, politics, and technology, and an individual's attitude on one topic is likely to be influenced or moderated by other related topics. For example, Myers and Bishop \cite{myers1970discussion} found that group discussions about racial attitudes not only reinforced participants' pre-existing racial prejudices, but also influenced their views on other social issues (e.g., education policy). The study demonstrated that when people engage in a group discussion about a particular issue, not only are attitudes about that issue polarized, but the effect may also spill over to other related issues.

To address the above issues, this paper proposes the MTOS framework for simulating the cognitive evolution process in a multi-topic environment. Each agent in MTOS is initialized with a unique role, including gender, age, educational background, and personal traits, and an initial opinion vector is set on multiple topics to reflect its multidimensional position. Agents establish heterogeneous connections within a scale-free network to simulate the diverse interaction structures among individuals in real-world societies. Agents are equipped with short-term memory to record daily interactions and long-term memory to capture historical cognition and contextual background, while incorporating a dynamic decay mechanism to adjust historical information. In each round, agent selects neighbors for exchanging opinions based on the similarity of multidimensional average belief values or the semantic matching mechanism of structured prompts. It then analyzes and reasons by integrating multi-topic beliefs and role attributes, updating opinions across various topics. The multi-topic recommendation mechanism dynamically adjusts an agent's attention bias toward different topics by simultaneously considering both group topic popularity and individual topic fatigue. This enables the agent to generate recommended topics aligned with its own state and environmental needs, fostering topic competition and enhancing the authenticity and coherence of interactive behavior. Within a multi-topic interaction framework, the agent integrates neighbors' multi-topic opinions through short-term memory and updates belief values by combining them with existing cognitive states stored in long-term memory. Opinions across different topics interact and propagate information via memory structures. The belief decay mechanism dynamically regulates the magnitude and speed of updates to simulate cognitive fatigue effects, enabling agents to generate multi-topic opinion evolution consistent with human cognitive characteristics. To evaluate the effectiveness of MTOS, we conducted systematic simulation experiments on typical network architectures. The study focused on comparing the evolution of group opinions under single-topic versus multi-topic conditions. It also examined the impact of different topic correlation combinations (strong positive, weak positive, none, strong negative, and weak negative) on the echo chamber effect. We compared the performance of MTOS with the existing SSF in terms of echo chamber related metrics. MTOS demonstrates more realistic simulation outcomes on both the neighbor-related index and global divergence metric. It not only reproduces the echo chamber phenomenon observed in the real world but also offers new insights into the role of topic-related patterns in group cognitive dynamics.

In summary, based on the above findings, we make the following contributions:

1. We propose the \textbf{MTOS} framework to model the evolution of group opinions across multiple topics, powered by LLMs.

2. We incorporate multiple strategies and dynamic topic recommendation algorithms to enhance the pragmatic consistency and social plausibility of simulated behaviors.

3. Through extensive simulation experiments, we systematically evaluate the impact of different models on the echo chamber effect. Results indicate that, compared to other models (e.g.,SSF), our approach better replicates the echo chamber phenomenon in multi-topic social environments, while providing a more interpretable tool for modeling the mechanisms and intervention pathways of the echo chamber effect.

\section{Related Work}
\subsection{Dynamics of Opinions Model}
In recent years, researchers have proposed various modeling approaches for multi-topic opinion evolution. Giulia De Pasquale et al.~\citep{DePasqualeValcher2022} developed two multidimensional opinion dynamics models based on the Hegselmann–Krause finite trust model. These models employ interaction rules based on average opinion and topic-by-topic similarity, respectively.They are used to characterize the dynamic evolution of individual opinions in multi-topic scenarios. Chen et al.~\citep{ChenZhangHuang2017} proposed a multi-topic evolution model based on the SIR framework that incorporates interference similarity between topics, systematically analyzing the moderating effect of topic relevance on information transmission probability. In recent research, Li et al.~\citep{LiLuoChu2025} further extended the finite trust model to a multidimensional opinion space. By introducing a weighted topic inconsistency function, they revealed the pivotal role of topic relevance in the clustering and evolution of multi-topic opinions. In multi-topic environments, prominent competition exists between information or topics. Existing research has characterized this competitive mechanism from various perspectives. Some studies analyze the competitive nature of topics through quantitative metrics and establish multi-topic communication models, revealing that a topic's popularity significantly influences user selection behavior~\citep{SunZhouGuan2016, chen2019multi}. Further research examining the relationship between users and topics has proposed methods for predicting topic popularity and tracking shifts in attention, providing both theoretical and empirical support for understanding user choice dynamics in multi-topic environments~\citep{yu2017, an2024}. Furthermore, Weng et al.~\citep{weng2012competition} investigated competitive phenomena in scenarios where multiple pieces of information coexist using a subject-based social network model. They found that a minority of information can spread virally, while the majority remains ignored, exhibiting significant heterogeneity in popularity and persistence. Multi-topic competition not only influences information popularity and user choice behavior but also serves as a key driver of public opinion evolution. However, existing research still lacks a simulation framework capable of systematically modeling the cognitive evolution of multiple topics at the level of linguistic interaction. Building upon the aforementioned theories, this paper leverages the generative capabilities of LLMs. We construct an opinion polarization evolution model that integrates multi-topic competition with semantic generation. This approach aims to more comprehensively capture the complex mechanisms underlying the dynamic evolution of opinions within multi-topic environments.

\subsection{LLM-based Opinion Propagation}
Integrating LLMs into social dynamics simulations is an emerging field of research that has yielded promising results. Yao et al.~\citep{Yao2025} proposed an algorithm (FDE-LLM) that integrates opinion dynamics equations with LLMs. By simulating the interaction process between opinion leaders and followers, this approach effectively captures the evolution dynamics of opinions under multi-role scenarios. It also significantly enhances simulation accuracy and efficiency. This demonstrates the application potential of LLMs in modeling complex social opinion evolution. Wang et al.~\citep{WangLiuYangChen2025} utilized LLM-driven agent interactions to replicate echo chambers and opinion polarization processes, proposing language-based intervention strategies to mitigate polarization. Hu et al.~\citep{Hu2025} designed multiple LLM-based agents incorporating diverse network architectures to simulate rumor propagation processes. Their work demonstrated the superiority of LLM agents in capturing individual behaviors and social interaction details, revealing the critical influence of network topology and agent behavior on rumor diffusion. These studies demonstrate that integrating LLMs with traditional models provides more authentic and interpretable approaches for social opinion research. However, while existing LLM-based studies have made significant progress in simulating opinion propagation and polarization within single-topic contexts, they have not sufficiently addressed the complex realities of social environments where multiple topics coexist and interact. This paper integrates an opinion dynamics model with the generative capabilities of LLMs. We construct an opinion evolution model that fuses multi-topic interaction with semantic generation. This approach effectively captures the interplay between topics and the dynamic evolution of opinions, offering new methodologies and perspectives for simulating multi-topic communication in social environments.

\section{Method}
\subsection{Overview}
\begin{figure}[htbp]
\centering
\includegraphics[width=0.85\textwidth]{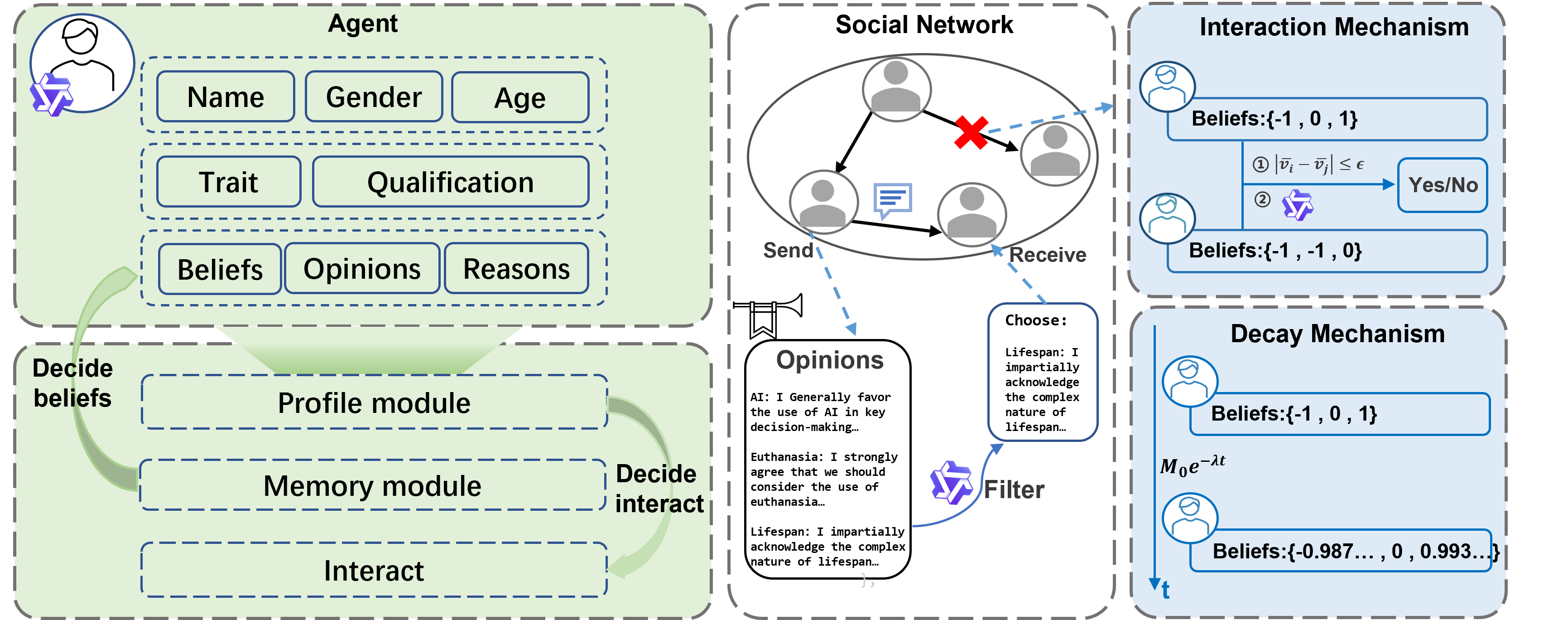}
\caption{The left panel displays user profiles, the middle section illustrates the dynamic topic selection mechanism within social networks and frameworks, while the right panel presents the interaction target selection mechanism and dynamic decay mechanism.}
\label{Fig1}
\end{figure}

We constructed a multi-topic simulation environment based on the \textbf{MTOS} framework. This environment consists of $N$ agents driven by LLMs, denoted as:
\begin{equation} \label{eq:agents}
A = \{ a_1, a_2, \dots, a_N \},
\end{equation}
where $A$ represents the set of all agents and $a_i$ denotes the $i$-th agent in the environment. The collection of topics is defined as:
\begin{equation} \label{eq:topics}
T = \{ T_1, T_2, \dots, T_K \},
\end{equation}
where $T$ denotes the set of all topics, and $T_k$ represents the $k$-th topic. During the initialization phase, agents establish connections within a scale-free network to reflect the heterogeneous interaction structures found in real-world societies. Each agent is assigned a unique social role and a set of initial belief values across multiple topics. The initial belief vector of agent $a_i$ is represented as:
\begin{equation} \label{eq:belief values}
\mathbf{v}_i^{(0)} = [v_{i,1}^{(0)}, v_{i,2}^{(0)}, \dots, v_{i,K}^{(0)}],
\end{equation}
where $v_{i,k}^{(0)} \in [-2, 2]$ denotes agent $a_i$'s initial opinion on topic $T_k$, with values ranging from strong opposition ($-2$) to strong support ($2$).During each simulation cycle (denoted as day $t$), agent $a_i$ selects a user for interaction either from its neighbors—based on the multidimensional Hegselmann–Krause (HK) model—or according to a recommendation strategy. Then, each pair of interacting agents randomly selects a topic from the intersection of their recommended topic sets for opinion exchange. After the interaction concludes, each agent integrates and reflects on the received information. It then updates its belief value on the corresponding topic through a dynamic decay mechanism, enabling the continuous and adaptive evolution of its perspective over time.The overall framework is shown in Figure.~\ref{Fig1}.

\subsection{Multi-topic Agent Interaction Mechanism}
To more authentically recreate the complex process of opinion interaction among users in multi-topic social environments, this paper proposes and implements two types of multi-topic agent interaction mechanisms. The first is a similarity filtering mechanism that integrates belief-value averaging algorithms with social network structures. The second is a multi-attribute semantic matching mechanism based on structured prompts.

\paragraph{Interaction Mechanism Based on the Multidimensional Hegselmann–Krause Model}
Considering the complex interaction behaviors of users in multi-topic environments, this study introduces a multi-topic user interaction mechanism that integrates the multidimensional Hegselmann–Krause (HK) model with social network structural features. Unlike traditional interaction rules based on single-topic opinions, this mechanism dynamically adjusts the selection of interaction neighbors among agents according to users’ overall positions in a multidimensional opinion space, aiming to more authentically reflect the diverse interaction patterns observed in real-world social networks. Specifically, the belief vector of each agent $a_i$ consists of elements $v_{i,k}$, where $v_{i,k}$ represents the agent's belief on topic $T_k$. To capture the overall opinion tendency of an agent across multiple topics, the average belief value is defined as:
\begin{equation} \label{eq:averge belief values}
\bar{v}_i = \frac{1}{K} \sum_{k=1}^{K} v_{i,k}.
\end{equation}
In the neighbor selection process, the threshold parameter $\epsilon$ defines the permissible range of belief similarity for interaction between agents, thereby controlling the scope of neighbor recommendations. The neighbor set of agent $a_i$ is defined as:
\begin{equation} \label{eq:neighbor}
N_i = \{ a_j \in M(i) \; | \; |\bar{v}_i - \bar{v}_j| \leq \epsilon \},
\end{equation}
where $M(i)$ denotes all adjacent neighbors of $a_i$ in the social network. Interaction between two agents occurs only when the difference in their average belief values satisfies the above condition. This mechanism effectively avoids interactions across excessively divergent opinions, thereby reflecting the principle of \textit{homophily} commonly observed in real-world social networks.

\begin{figure}[htbp]
\centering
\includegraphics[width=0.8\textwidth]{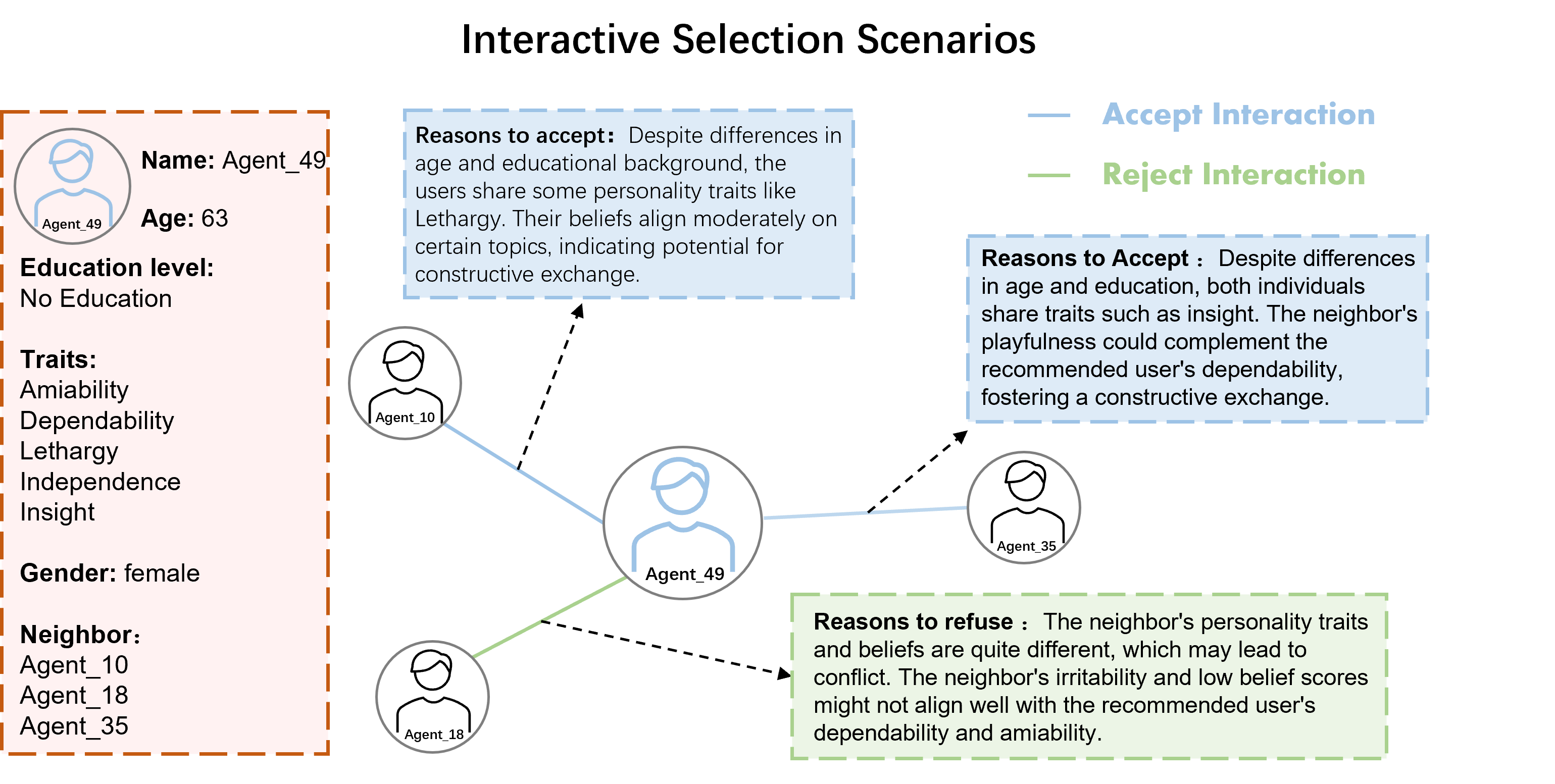}
\caption{This mechanism thus illustrates the decision-making logic of agents when interacting with neighbors in social networks, including the specific reasons for accepting or rejecting.}
\label{Fig2}
\end{figure}

\paragraph{Semantic Matching Screening Mechanism Based on Structured Prompts}
To further enhance the personalization and semantic alignment of interactive recommendations, this paper designs a semantically driven filtering mechanism that integrates multi-attribute information with multi-topic belief matching. Leveraging the reasoning capabilities of LLMs, it enables semantic hierarchical evaluation of candidate neighbors. Within this mechanism, the system automatically constructs a structured prompt for each candidate interaction object for the language model to evaluate. The prompts organize information in a structured template format, covering:
\begin{itemize}
    \item \textbf{Agent's fundamental attributes:} such as age, gender, education level, and personality traits.
    \item \textbf{Belief distribution:} the agent’s belief values across multiple topics; for example, the agent’s beliefs on topics A, B, and C might be $[1, 0, -1]$, indicating support, neutral, and oppose stances, respectively.
\end{itemize}
Based on the input information above, a structured semantic description template is constructed and passed to the LLM for reasoning and decision-making. The model is required to produce output in a predefined JSON format, returning either ``\texttt{yes}'' or ``\texttt{no}''. This indicate whether the recommended entity should interact with this neighbor. This structured output ensures the clarity and consistency of interaction decisions. The reasons provided by LLMs for whether adjacent users interact with each other are shown in Figure~\ref{Fig2}. The filtering process iteratively evaluates all neighbors in the candidate set, performing personalized neighbor selection within a high-dimensional feature space by semantically matching the combined attributes and multi-topic belief differences between agents. Finally, multi-topic agent interaction mechanism identifies a set of neighbors exhibiting high semantic affinity and multidimensional attribute similarity, thereby constructing a more precise and multi-topic-coupled interaction network structure.

\subsection{Multi-Topic Recommendation Mechanism}
To simulate topic competition in multi-topic environments, this study designs a dynamic topic recommendation mechanism that assigns or recommends topics for agents to participate in based on their individual attributes and historical behaviors. In each interaction round, an agent receives only a single topic recommended by the system for expressing opinions and engaging in communication, thereby controlling the precision and diversity of topic transmission. The mechanism first aggregates topic selection data from all agents' historical interactions. Through statistical analysis, it calculates the overall popularity of each candidate topic, reflecting its prevalence within the group. Meanwhile, for individual agents, an agent-specific topic fatigue metric is computed based on their historical topic selection frequency, characterizing both the agent's exposure to specific topics and the potential diminishing effect of interest. During the recommendation process, the system comprehensively considers the agent's fundamental demographic attributes (including age, gender, personality traits, and education level), long-term memory content, current conversational context, current topic set, and the aforementioned statistical metrics. By constructing a language model prompt tailored to the agent and inputting it into a large-scale pre-trained LLMs, multi-topic recommendation mechanism selects the most suitable topic recommendation corresponding to the agent's current state.
\begin{figure}[htbp]
\centering
\includegraphics[width=1.0\textwidth]{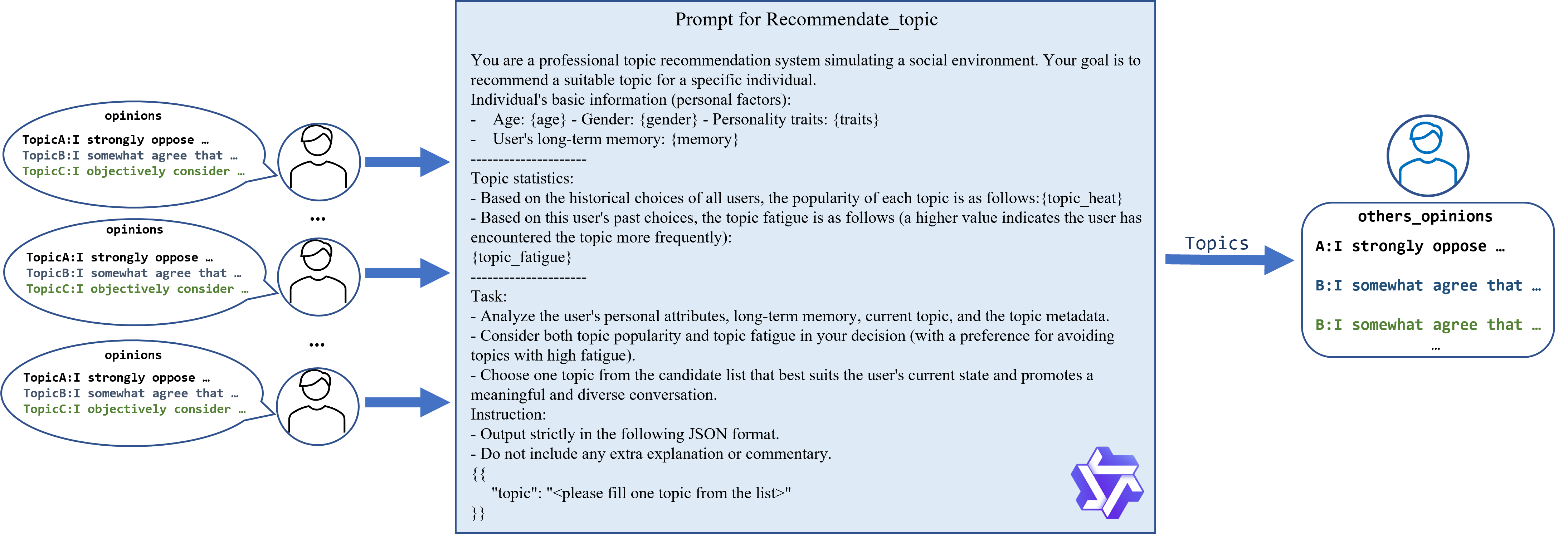}
\caption{Topic Selection Mechanism.}
\label{Fig3}
\end{figure}
Let the candidate topic set be: \begin{equation} \label{eq:candidate topic}
\mathcal{T} = \{T_1, T_2, \dots, T_K\}, 
\end{equation} and the agent set be A. Each agent $a_i$ possesses a set of basic attributes $p_i$, including age, gender, personality traits, and educational level.
The system first aggregates historical topic selection data from all agents to obtain the universal set:
\begin{equation} \label{eq:H}
H = \bigcup_{i=1}^{N} H_i,
\end{equation}
where $H_i$ denotes the historical topic selection sequence of agent $a_i$. Based on $H$, the overall popularity (heat) of each topic $T_k$ is calculated as the relative frequency of the topic across all interactions:
\begin{equation} \label{eq:heat}
\text{Heat}(T_k) = \frac{|\{h \in H : h = T_k\}|}{|H|}.
\end{equation}
To model cognitive fatigue from repeated exposure to the same topic, we introduce an agent-specific \textbf{topic fatigue metric}. Let $H_i$ denote the historical interaction topics of agent $a_i$, and define the topic selection rate (TSR) for topic $T_k$ as:
\begin{equation} \label{eq:TSR}
\text{TSR}_i(T_k) = \frac{|\{h \in H_i : h = T_k\}|}{|H_i|}.
\end{equation}
The fatigue level is then calculated via a parameterized exponential mapping function:
\begin{equation} \label{eq:Fatigue}
\text{Fatigue}_i(T_k) = \frac{\exp(b \cdot \text{TSR}_i(T_k)) - 1}{\exp(b) - 1},
\end{equation}
where $b>0$ controls the steepness of fatigue intensity. This function is monotonic and normalized to $[0,1]$, such that a high $\text{TSR}_i(T_k)$ leads to fatigue approaching 1 (low willingness to interact), while a low $\text{TSR}_i(T_k)$ corresponds to a high willingness to engage.
Based on the agent's individual attributes $p_i$, long-term memory $M_i^l$, topic set $\mathcal{T}$, topic heat $\{\text{Heat}(T_k)\}$, and fatigue levels $\{\text{Fatigue}_i(T_k)\}$, we construct the input prompt for the LLMs:
\begin{equation} \label{eq:X}
X_i = f(p_i, M_i^l, \mathcal{T}, \{\text{Heat}(T_k)\}, \{\text{Fatigue}_i(T_k)\}).
\end{equation}
As shown in Figure~\ref{Fig3}, we present the prompt for topic recommendation.This prompt is provided to the LLMs in a structured format to generate topic recommendations aligned with the current interaction context and agent state:
\begin{equation} \label{eq:LLM}
T_i^{\text{rec}} = \text{LLM}(X_i).
\end{equation}
This mechanism thus effectively balances group topic popularity trends with individual dynamic interests, simulates competition for topic, promotes personalized and diverse topic selection, and enhances the realism and coherence of multi-topic interactions. Leveraging the cognitive reasoning capabilities of LLMs, this approach achieves complex, dynamic decision-making based on linguistic context, overcoming the limitations of traditional rule-based recommendation methods.

\begin{figure}[htbp]
\centering
\includegraphics[width=0.8\textwidth]{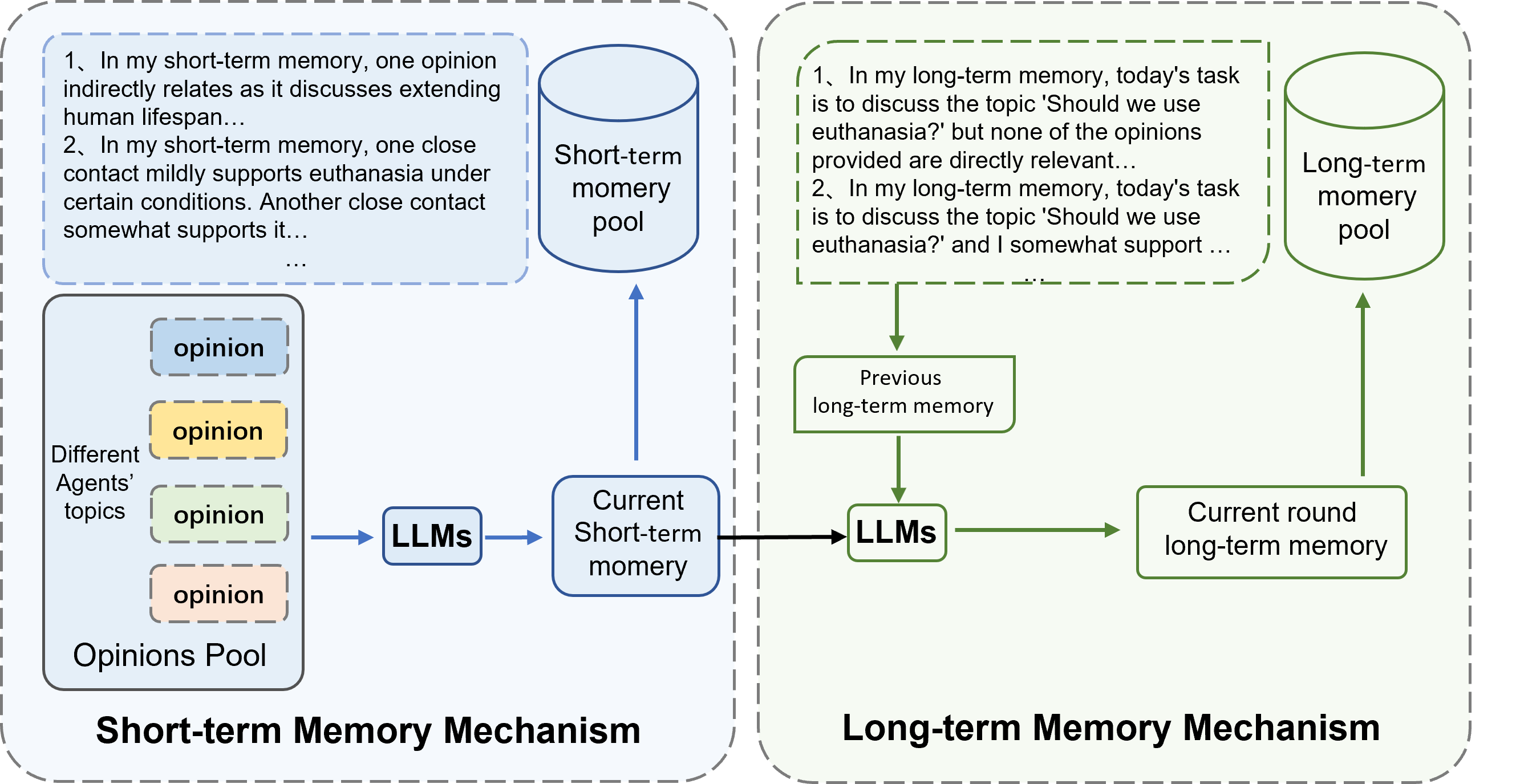}
\caption{Dual-layer memory structure.}
\label{Fig4}
\end{figure}

\subsection{Multi-Topic Belief Update Mechanism}
After introducing the interaction mechanism, we describe how agents update their multi-topic beliefs within this framework. Each agent is randomly assigned a role attribute, which includes personal information, belief values, and other relevant characteristics. In this model, we employ a dual-layer memory architecture:
\begin{itemize}
    \item \textbf{Short-term memory:} Aggregates recent information from daily interactions.  
    \item \textbf{Long-term memory:} Accumulates historical knowledge and contextual information over time.
\end{itemize}
Figure~\ref{Fig4} shows the decomposition of dual-layer memory and a portion of the content currently stored in the memory pool. Based on this architecture, we introduce a belief decay mechanism to reflect the cognitive fatigue effect that occurs when individuals focus on a particular topic over an extended period. Let $\lambda > 0$ denote the decay coefficient. The update of an agent's belief value $v(t)$ at time $t$ is defined as:
\begin{equation} \label{eq:decay}
v(t+1) = v(t) \cdot e^{-\lambda \cdot |v(t)|}.
\end{equation}
Specifically, during daily interactions, each agent receives a diverse collection of topic-related opinions from multiple neighbors. These heterogeneous inputs are semantically compressed and integrated through the short-term memory module. Subsequently, by combining these with existing cognitive states stored in long-term memory, the agent updates its belief values based on the most recent interaction information.  Opinions across different topics interact and transmit information through this memory structure, thereby reflecting cross-topic cognitive influence in a multi-topic environment. The decay mechanism dynamically adjusts the magnitude and speed of belief changes, simulating the adaptive weakening of cognitive updating under sustained cognitive load. Ultimately, incorporating the decay mechanism and cross-topic influence, each agent generates updated opinions based on the adjusted belief values, achieving a multi-topic opinion evolution process that aligns more closely with human cognitive characteristics.

\section{Experiments}
\subsection{Experiments Setup}
This study constructs a multi-topic social simulation framework, \textbf{MTOS}. Both agents and their environment are implemented using the Python library Mesa \cite{KazilMasadCrooks2020}. To avoid bias stemming from names, each agent is identified solely by a number, with gender and age randomly assigned within specified ranges (age range: 18 to 64 years). The simulation comprises 50 agents, which representing a notable increase compared to previous LLM-based simulation studies \cite{Liu2024a, Chuang2024}, thereby enhancing the social complexity of the model. Agents' personality traits are randomly assigned based on the widely used Big Five personality model \cite{BarrickMount1991}, with each dimension having an equal probability of exhibiting positive or negative characteristics. For language modeling, this study employs the locally deployed large language model Qwen-2.5 (7B) as the cognitive engine for each agent, enhancing reasoning consistency and controllability of simulation outcomes. All simulations are conducted in a local environment equipped with NVIDIA GeForce RTX 4070 graphics cards, to ensuring efficient model inference and resource scheduling.To better reflect multi-topic communication characteristics in real social scenarios, we design topics with different semantic categories and emotional tones. Each agent gradually accumulates opinions, memories, and updates around a single topic over multiple rounds of interaction, thereby simulating the opinion evolution mechanism under multi-topic conditions. The parameters used in the simulation are summarized in Table~\ref{tab:parameters}.

\begin{table}[H]
\centering
\caption{Parameters used in the experiment}
\label{tab:parameters}
\begin{tabular}{l l}
\toprule
\textbf{Parameter} & \textbf{Numerical Value} \\
\midrule
Running Rounds & 30 \\
Number of Agents & 50 \\
Age of Agents & 18--64 \\
Personality of Agents & Openness, Conscientiousness, Extraversion, Agreeableness, Neuroticism \\
Network Infrastructure & scale-free network \\
Gen\_temperature & 0.5 \\
Tolerance thresholds for neighbor selection & 0.1 \\
Belief value similarity threshold $\epsilon$ & 0.1 \\
Seed & 50 \\
Fatigue sensitivity $b$ & 5.0 \\
Decay factor $\lambda$ & 0.01 \\
\bottomrule
\end{tabular}
\end{table}

\subsection{Indicators}
In order to comprehensively assess the performance of the model in simulating the echo chamber effect, this paper designs two types of assessment metrics to characterize both the dynamics of the system's evolution over time and the comparative effects of different mechanisms at the overall level.

\paragraph{Continuous Evolution Indicator}
This class of metrics is designed to quantify the evolution of individual opinions over the simulation cycles, aiming to reveal trends in the formation and development of echo chamber structures\cite{WangLiuYangChen2025}. It consists of the following components:
\paragraph{Neighbor Correlation Index(NCI)} 
The Neighbor Correlation Index (NCI) quantifies the degree of similarity between an individual's belief value and the average belief value of their neighborhood, reflecting trends in opinion congruence at the local level. Higher NCI values generally indicate stronger local coherence structures within social networks.
Formally, for an agent $a_i$ and its neighbors $a_j \in N_i$, we determine whether the difference between their beliefs on a topic $t$ is less than a threshold $\epsilon$. The individual-level NCI for agent $a_i$ is computed as:
\begin{equation} \label{eq:NCI_i}
\text{NCI}_i^{\text{threshold}} = \frac{1}{|N_i|} \sum_{a_j \in N_i} \mathbf{1}\left(|v_{i,t} - v_{j,t}| < \epsilon \right),
\end{equation}
where $v_{i,t}$ and $v_{j,t}$ denote the belief values of agents $a_i$ and $a_j$ on topic $t$, and $\mathbf{1}(\cdot)$ is the indicator function that outputs 1 if the condition is satisfied and 0 otherwise.
The global NCI across all $N$ agents is then obtained by averaging over all individuals:
\begin{equation} \label{eq:NCI}
\text{NCI}^{\text{threshold}} = \frac{1}{N} \sum_{i=1}^{N} \text{NCI}_i^{\text{threshold}}.
\end{equation}
Higher values of $\text{NCI}^{\text{threshold}}$ indicate stronger local coherence and more pronounced echo chamber tendencies.

\paragraph{Echo Chamber Index(ECI)} 
The Echo Chamber Index (ECI) quantifies the degree of similarity in topic opinions between an individual and their neighbors, characterizing the strength of the echo chamber effect within a local network. For a given topic $t$, the ECI measures the average similarity of opinions between each agent $a_i$ and its neighbors $a_j \in N_i$, and then averages across all agents to obtain the network-level index. The belief difference is normalized to the range $[0,1]$ by dividing by 4. Values closer to 1 indicate greater local consensus and a more pronounced echo chamber effect. Formally, the ECI is defined as:
\begin{equation} \label{eq:ECI}
\text{ECI} = \frac{1}{N} \sum_{i=1}^{N} \frac{1}{|N_i|} \sum_{a_j \in N_i} \left( 1 - \frac{|v_{i,t} - v_{j,t}|}{4} \right),
\end{equation}
where $v_{i,t}$ and $v_{j,t}$ denote the belief values of agents $a_i$ and $a_j$ on topic $t$, respectively.

\paragraph{Overall Comparison Metrics}
To assess the impact of different model mechanisms on population evolution, this study records the change in aggregate indicators ($\Delta$ indicator) after each simulation round to evaluate their influence on the echo chamber effect.

\paragraph{Polarization (P)} 
The degree of opinion polarization within a group on a specific topic is measured by the variance of all individual belief values. Higher values indicate greater dispersion of group opinions and more severe polarization. The polarization of topic $t$ at time $t$ is calculated as:
\begin{equation} \label{eq:P}
P_t = \frac{1}{N} \sum_{i=1}^{N} \left( v_{i,t} - \frac{1}{N} \sum_{j=1}^{N} v_{j,t} \right)^2,
\end{equation}
and the change in polarization relative to the initial state is:
\begin{equation} \label{eq:Pt}
\Delta P = P_t - P_0.
\end{equation}
A positive $\Delta P$ indicates increased overall polarization, greater group opinion divergence, and a strengthened echo chamber effect. Conversely, a negative $\Delta P$ suggests decreased polarization, which mitigates opinion divergence, though excessively negative values may lead to over-homogenization of the group.

\paragraph{Global Disagreement (GD)} 
Global Disagreement quantifies the divergence of opinions between an individual and their neighbors across all topics, capturing latent conflicts within the group. For agent $a_i$ with neighbors $a_j \in N_i$, GD is defined as:
\begin{equation} \label{eq:GD}
\text{GD}_t = \frac{1}{2N} \sum_{a_i \in A} \frac{1}{|N_i|} 
\sum_{a_j \in N_i} w_{ij} (v_{i,t} - v_{j,t})^2,
\end{equation}
where $v_{i,t}$ denotes the opinion of agent $a_i$ on topic $T_t$, 
and $w_{ij}$ represents the weight of the connection between agents. 
The change in disagreement for topic $T_t$ over time is given by:
\begin{equation} \label{eq:GD1}
\Delta \text{GD} = \text{GD}_t - \text{GD}_0.
\end{equation}
A positive $\Delta \text{GD}$ indicates growing opinion divergence among neighbors, heightened local conflicts, and increased potential for group fragmentation. A negative $\Delta \text{GD}$ signifies greater consensus among neighbors, though excessively negative values may foster local echo chambers.

\paragraph{Mean Neighbor Correlation Index (Mean NCI)} 
The Mean Neighbor Correlation Index (Mean NCI) is computed across all individuals to measure the strength of local consistency throughout the social network. It quantifies the correlation between each agent's belief and the average belief of their neighbors.  The Mean NCI at time $t$ is calculated as:
\begin{equation} \label{eq:Mean NCI}
\text{Mean NCI}_t = \text{PearsonCorr}\Big( \{v_{i,t}\}_{a_i \in A}, \{\bar{v}_{N_i,t}\}_{a_i \in A} \Big),
\end{equation}
where $v_{i,t}$ is the belief of agent $a_i$ on topic $t$, and $\bar{v}_{N_i,t}$ is the average belief of $a_i$'s neighbors.  The change in Mean NCI relative to the initial state is defined as:
\begin{equation} \label{eq:Mean NCI1}
\Delta \text{Mean NCI} = \text{Mean NCI}_t - \text{Mean NCI}_0.
\end{equation}
A higher $\Delta \text{NCI}$ indicates increased local consistency and a stronger tendency toward echo chambers, which can reduce information diversity. Conversely, a lower $\Delta \text{NCI}$ signifies reduced local homogeneity and a mitigation of the echo chamber effect.

\begin{figure}[htbp]
\centering
\begin{subfigure}[b]{0.48\textwidth}
    \centering
    \includegraphics[width=\textwidth]{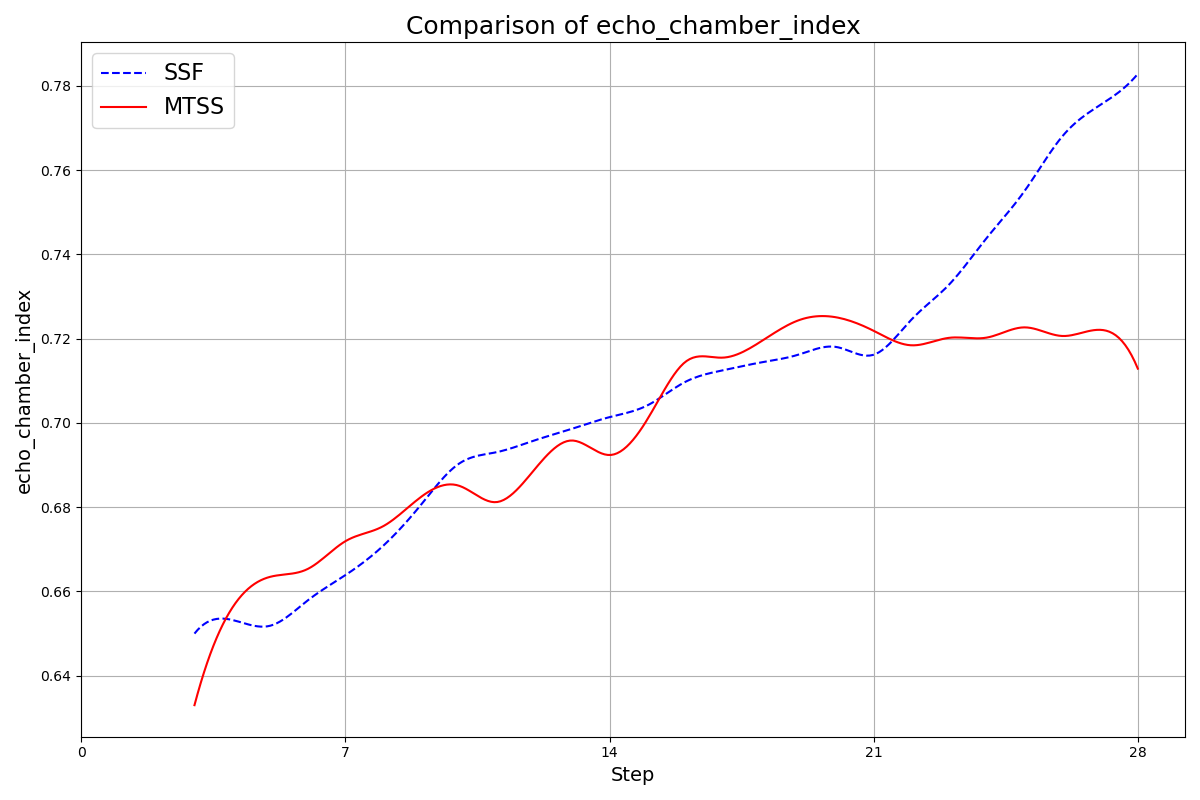}
    \caption{}
    \label{fig5a}
\end{subfigure}
\hfill
\begin{subfigure}[b]{0.48\textwidth}
    \centering
    \includegraphics[width=\textwidth]{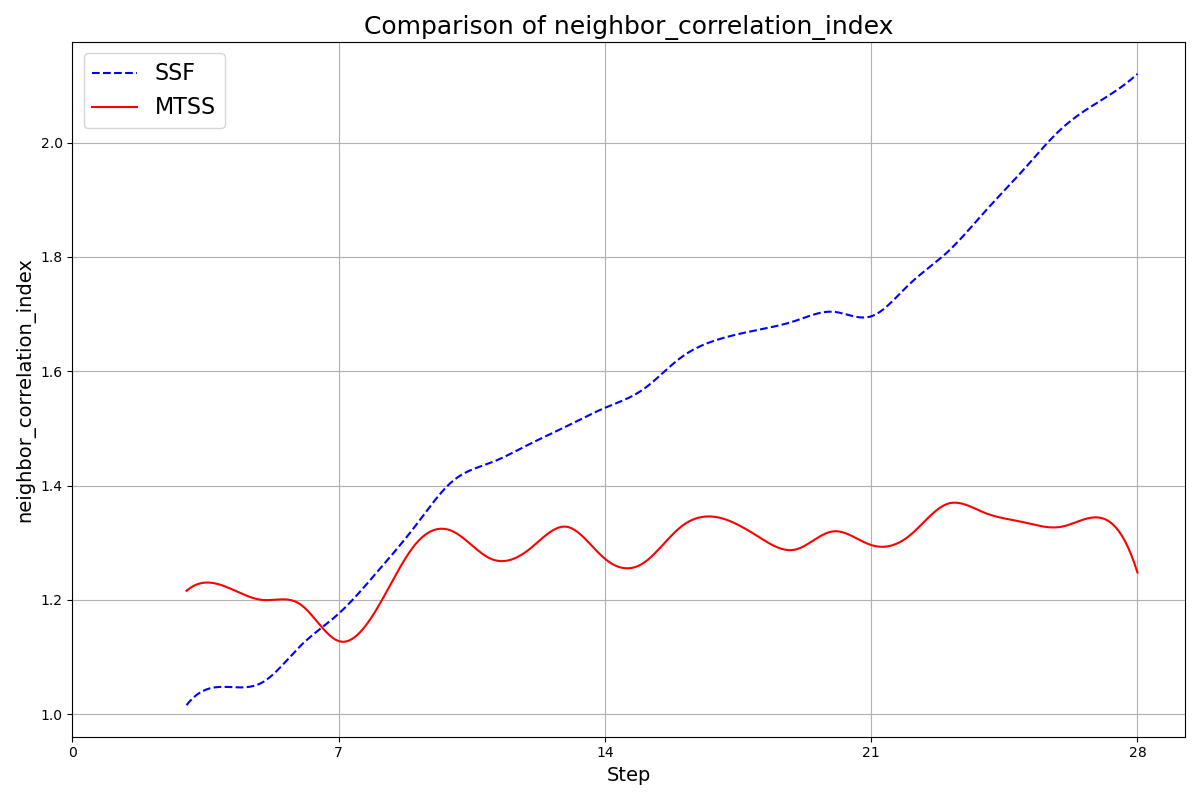}
    \caption{}
    \label{fig5b}
\end{subfigure}
\caption{Experimental results: (a) Echo Chamber Index; (b) Neighbor Correlation Index.}
\label{fig5}
\end{figure}

\subsection{Main Results}
\paragraph{Single-Topic vs. Multi-Topic}

In this study, to explore how information interaction in multi-topic environments influences group polarization and echo chamber effects, we designed two comparative experimental settings:\begin{enumerate}
    \item \textbf{Single-topic setting:} Agents update their opinions only around one core topic.
    \item \textbf{Multi-topic setting:} Agents hold opinions on multiple topics and simultaneously interact and update their views with their neighbors.\end{enumerate}
To examine the ``attention dispersion'' effect, we deliberately selected mutually unrelated topics, so that agents’ cognitive resources could not focus on a single topic, allowing us to observe the influence of cross-topic interactions on opinion formation. Due to the lack of correlation between topics, this dispersion can reduce the speed and intensity of opinion reinforcement on any single topic, which thereby mitigating the clustering of extreme views. In the experiments, we observed that opinions tended to be more diverse in the multi-topic environment, and polarization was reduced. This supports the notion that ``topic diversity'' affects the cognitive ecology of groups~\citep{Chen2021}.

\begin{figure}[htbp]
\centering
\includegraphics[width=1.0\textwidth]{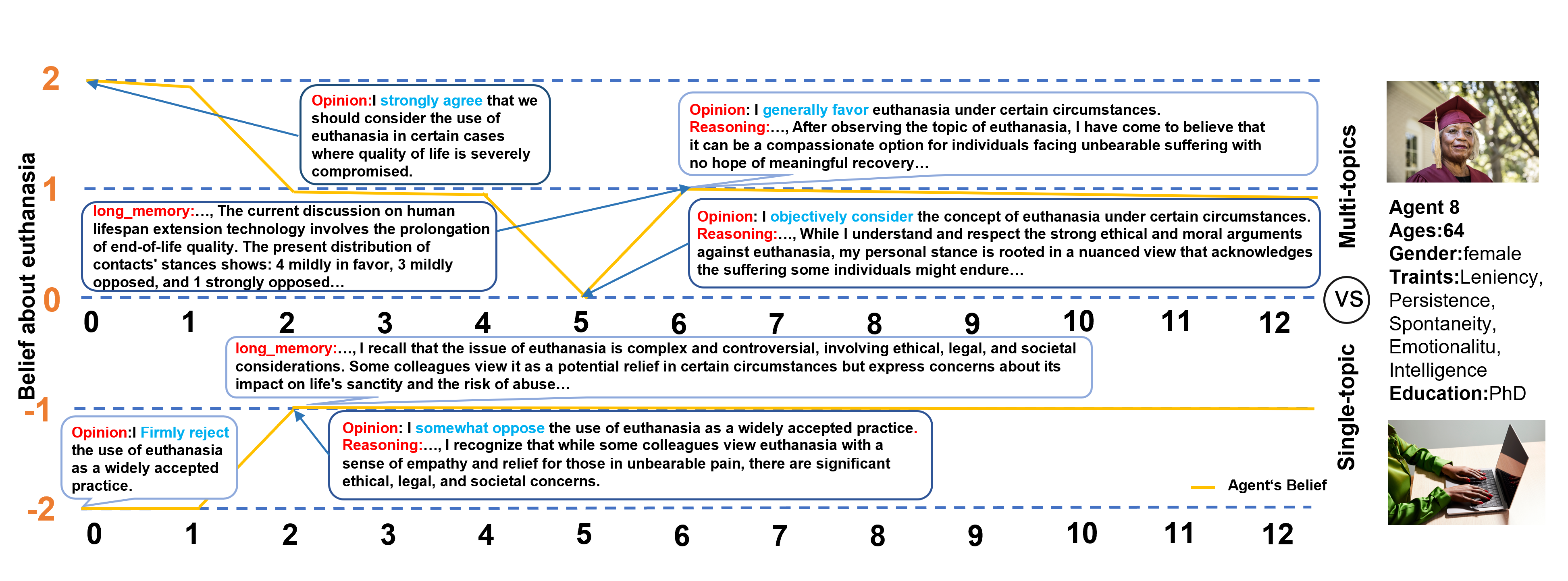}
\caption{The changes in agents’ belief values and their underlying reasons in single-topic and multi-topic experiments.}
\label{Fig6}
\end{figure}

Figure~\ref{fig5} shows the variations of the Echo Chamber Index and Neighbor Correlation Index under single-topic (SSF) and multi-topic (MTOS) environments. The results indicate that, in the single-topic environment, both the Echo Chamber Index and Neighbor Correlation Index exhibit a continuous upward trend, suggesting that individual opinions gradually converge, the echo chamber effect accumulates over time, and the correlation of opinions among neighbors increases, leading to a progressive intensification of local homogeneity. In contrast, the indices in the multi-topic environment are consistently lower than those in the single-topic environment, indicating that the introduction of unrelated topics in the multi-topic setting disperses individuals’ attention and effectively slows down the accumulation of opinion homogeneity. Overall, these findings suggest that the single-topic environment (SSF) facilitates the emergence of echo chamber effects at both global and local levels, whereas the multi-topic environment (MTOS) mitigates the accumulation of opinion homogeneity by diverting attention through unrelated topics, thereby alleviating the development of echo chambers.In Figure~\ref{Fig6}, we show the opinion changes of the same agent in single-topic and multi-topic environments, along with the underlying causes, agent profile, and selected memory fragments.

\begin{figure}[htbp]
\centering
\begin{subfigure}[b]{0.48\textwidth}
    \centering
    \includegraphics[width=\textwidth]{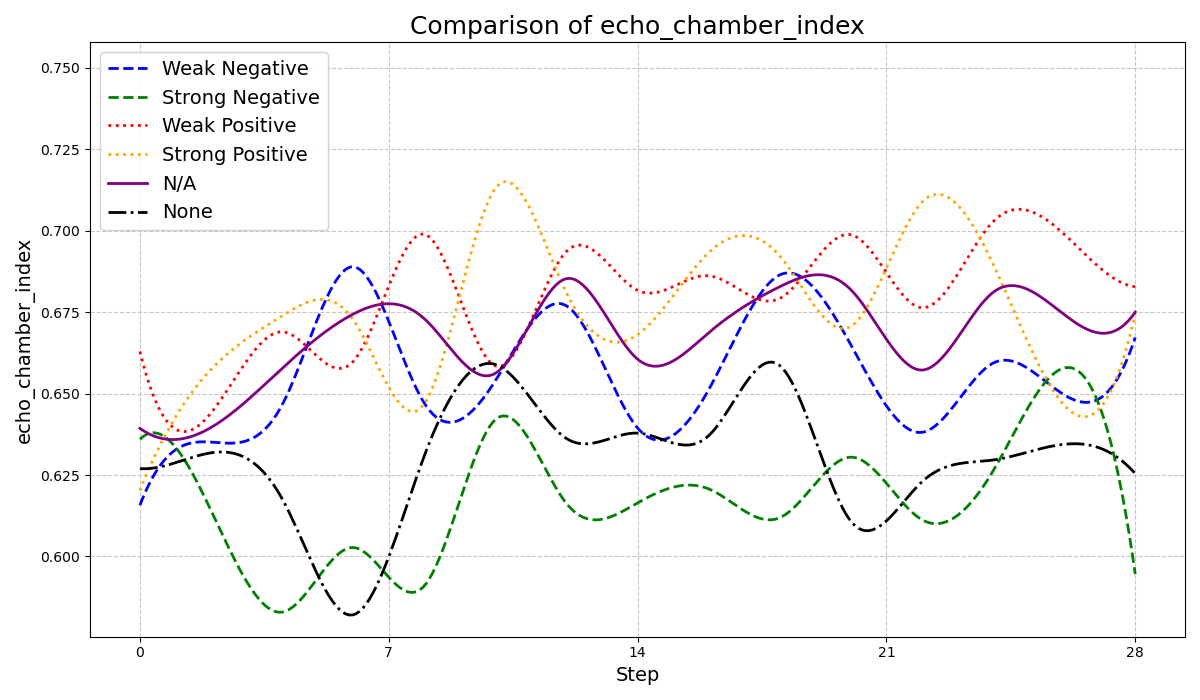}
    \caption{}
    \label{fig7a}
\end{subfigure}
\hfill
\begin{subfigure}[b]{0.48\textwidth}
    \centering
    \includegraphics[width=\textwidth]{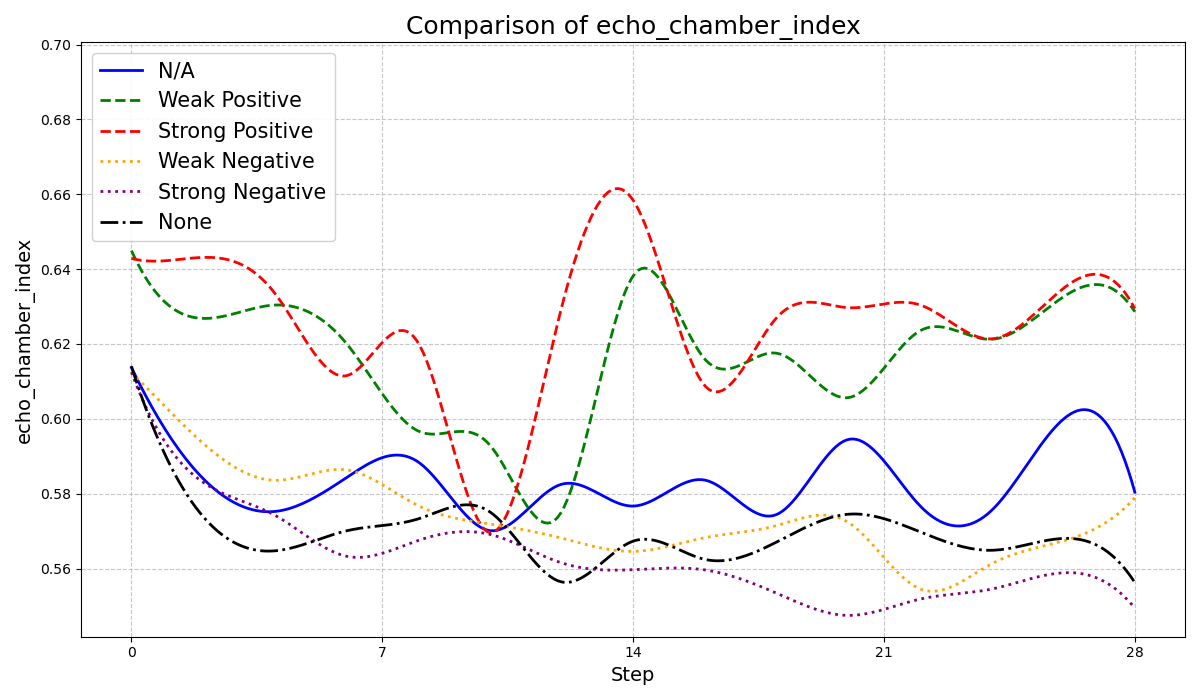}
    \caption{}
    \label{fig7b}
\end{subfigure}
\caption{Comparison of Echo Chamber Index under two interaction mechanisms: 
(a) Multi-dimensional Hegselmann-Krause  model; 
(b) Semantic Matching Screening Mechanism based on Structured Prompts.}
\label{fig7}
\end{figure}

\paragraph{Inter-Topic Correlation in Multi-Topic Experiments}
To investigate the influence of inter-topic correlation on the evolution of group opinions in a multi-topic system, this study designs six correlation scenarios in MTOS: 
\begin{itemize}
    \item \textbf{Single Topic Only (N/A):} Only the main topic is retained, without influence from other topics.
    \item \textbf{Strong Positive Relevance:} The new topic is highly aligned with the main topic in terms of values or positions.
    \item \textbf{Weak Positive Relevance:} The new topic shows a certain degree of support or indirect alignment with the main topic in values or positions.
    \item \textbf{Strong Negative Relevance:} The new topic is clearly opposed to the main topic in values or positions.
    \item \textbf{Weak Negative Relevance:} The new topic exhibits slight or indirect differences from the main topic in values or positions.
    \item \textbf{Irrelevant (None):} The new topic has no significant connection with the main topic in terms of beliefs or affective dimensions.
\end{itemize}
The experimental results indicate that, whether using the multi-dimensional Hegselmann-Krause (HK) model or the semantic matching screening mechanism based on structured prompts. Different topic correlation scenarios have significant effects on the evolution of group opinions. As observed in Figure~\ref{fig7}, strongly positively correlated topics reinforce existing consensus, leading to rapid convergence of opinions within the group; weakly positively correlated topics support existing positions to some extent, but the aggregation effect is relatively mild. In contrast, strongly negatively correlated scenarios introduce clear opinion conflicts, promoting diversification and opinion heterogeneity. Weakly negatively correlated topics slightly weaken the aggregation trend through minor opinion differences. By comparison, introducing unrelated topics disrupts cognitive focus on the main topic, dispersing individual attention across multiple dimensions, and thereby enhancing overall opinion diversity in the system. Overall, the strength and direction of topic correlations determine whether group opinions tend toward convergence or divergence. They also influence the formation and mitigation of echo chamber effects to varying degrees.

\subsection{Ablation Study}
\begin{figure}[htbp]
\centering
\begin{subfigure}[b]{0.48\textwidth}
    \centering
    \includegraphics[width=\textwidth]{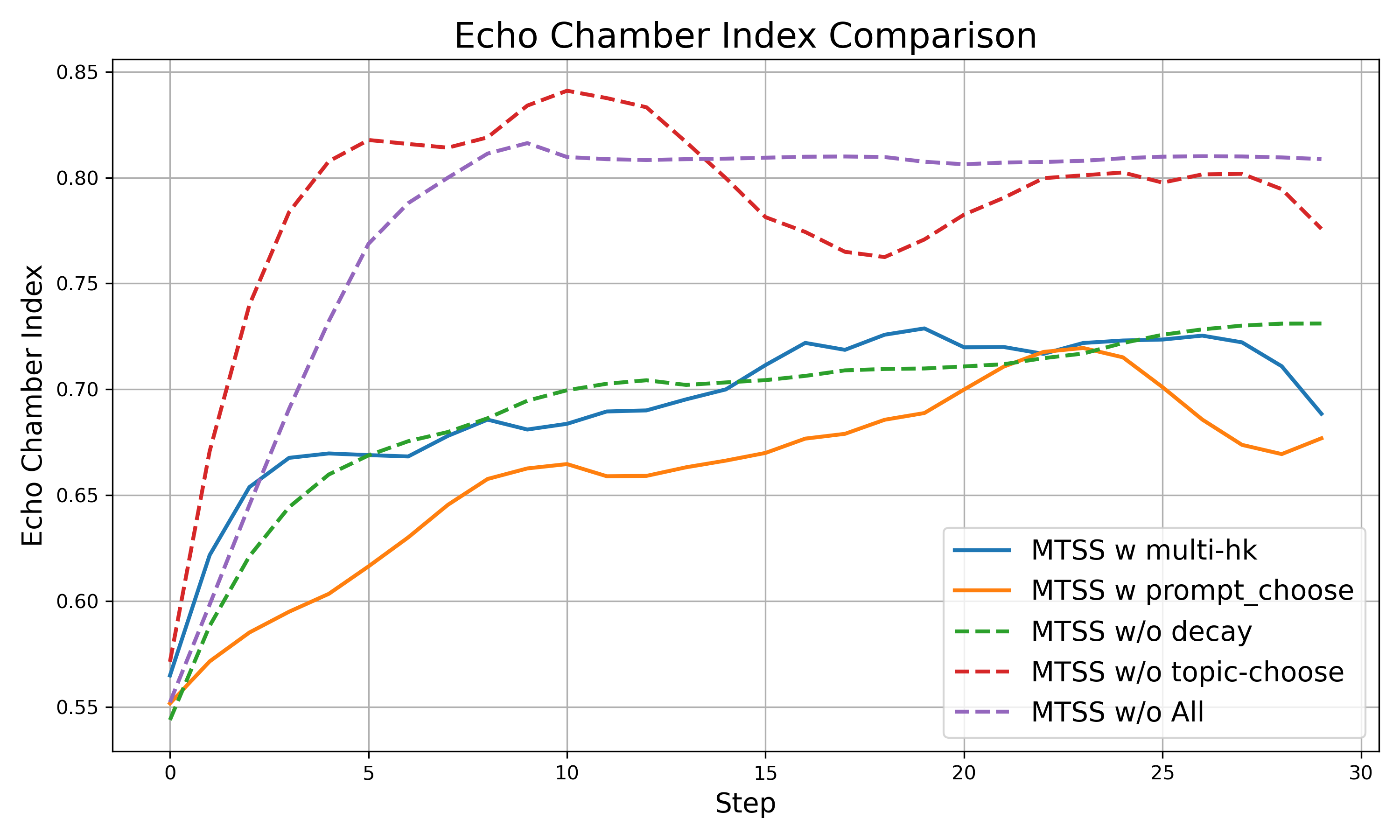}
    \caption{}
    \label{fig8a}
\end{subfigure}
\hfill
\begin{subfigure}[b]{0.48\textwidth}
    \centering
    \includegraphics[width=\textwidth]{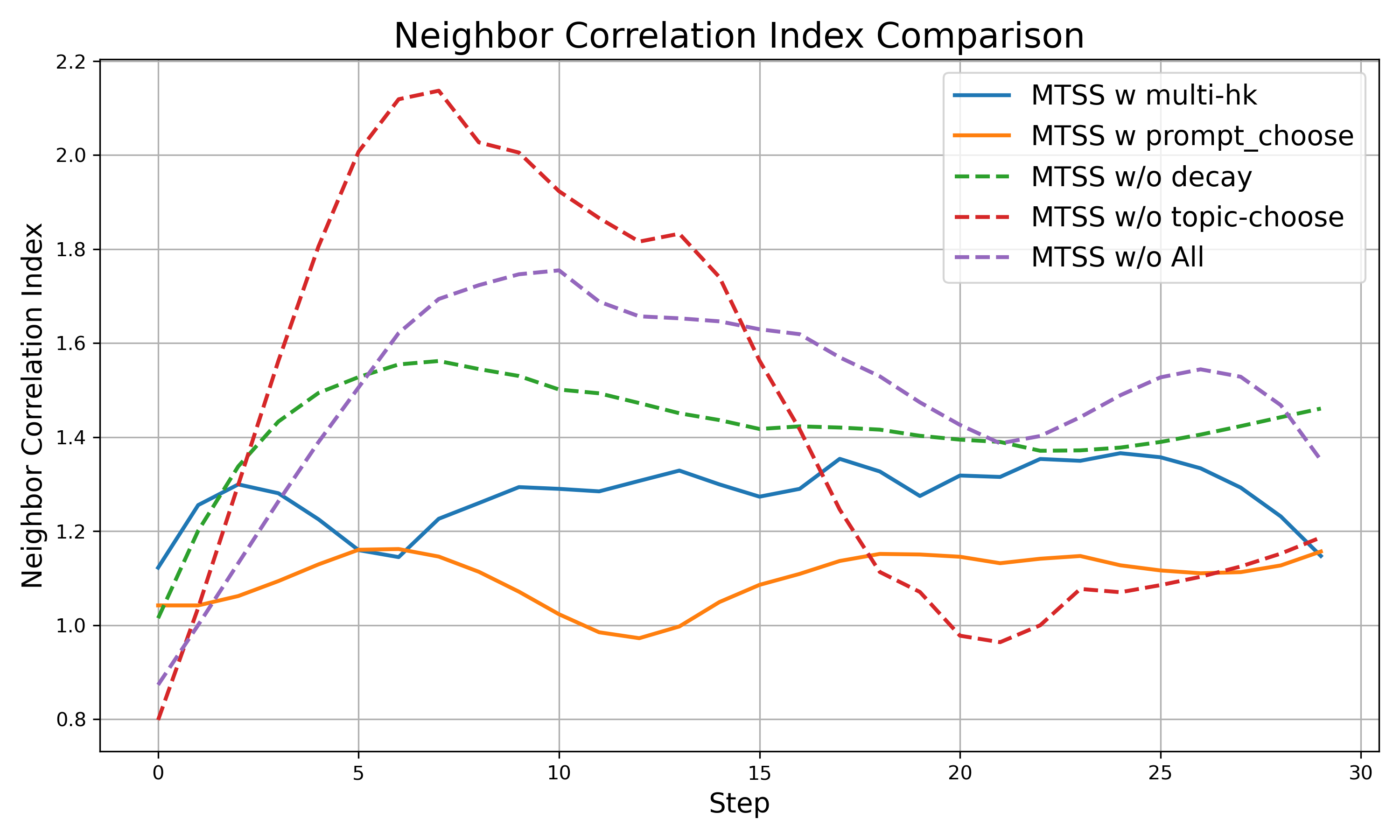}
    \caption{}
    \label{fig8b}
\end{subfigure}
\caption{Experimental results: (a) Echo Chamber Index; (b) Neighbor Correlation Index.}
\label{fig8}
\end{figure}

To examine the impact of key mechanisms in MTOS on echo chamber effects, this study conducted ablation experiments.In the ablation experiments, the multi-topic environment used introduces only mutually unrelated topics.The experiments systematically removed or replaced three core mechanisms: the multi-topic belief decay mechanism, multi-topic interaction mechanism, and the multi-topic recommendation mechanism, and quantified their effects using indicators such as the echo chamber index and neighbor correlation index (as shown in Figure~\ref{fig8}).Experimental results demonstrate that the proposed MTOS framework (whether employing the Multi-dimensional Hegselmann-Krause  model or the semantic Matching Screening Mechanism based on Structured Prompts) demonstrates strong effectiveness in mitigating the echo chamber effect. Compared to MTOS without multi-topic belief decay, MTOS without multi-topic recommendation, and MTOS without any mechanism, the model incorporating the neighbor selection mechanism shows significant improvement in local echo chamber metrics, with a markedly reduced Neighbor Correlation Index (NCI). This demonstrates that the complete MTOS framework effectively enhances information diversity and reduces the tendency toward excessive opinion clustering. Although the MTOS versions with HK model and semantic matching mechanism exhibit slight performance variations, both outperform the version without the mechanism. This validates the effectiveness of the proposed MTOS framework in maintaining community heterogeneity and mitigating the local echo chamber effect when incorporating unrelated topics.

We compared the changes in the Overall Comparison Metrics among SSF, MTOS, and MTOS variants with specific mechanisms removed to evaluate the mitigation of echo chamber effects in a multi-topic environment with unrelated topics. Table~\ref{tab:model_comparison} summarizes the variations in three indicators --- $\Delta$Polarization ($\Delta P$), $\Delta$Global Disagreement ($\Delta GD$), and $\Delta$Mean Neighbors Correlation ($\Delta$Mean NCI$)$ --- across different models. The overall results show that the complete MTOS framework achieves the best balance among the three metrics: both $\Delta P$ and $\Delta GD$ moderately decrease, indicating that it reduces polarization and global disagreement without causing excessive opinion convergence; meanwhile, $\Delta$NCI remains relatively low, suggesting that local homogeneity is effectively suppressed and group-level opinion heterogeneity is maintained. These findings demonstrate that MTOS enables information to spread and interact within a multidimensional issue space, enhancing system robustness and diversity. This alleviates echo chamber effects while maintaining openness and pluralism in group discussions.

\begin{table}[htbp]
\centering
\caption{Comparison of different models on echo chamber metrics.}
\begin{tabular}{lccc}
\toprule
Model & $\Delta$ Polarization $\downarrow$ & $\Delta$ Global Disagreement $\uparrow$ & $\Delta$ Neighbors Correlation $\downarrow$ \\
\midrule
SSF & -1.5700 & -1.9082 & 0.6043 \\
MTOS & -1.1855 & -1.5775 & \text{0.4791} \\
MTOS (prompt choose) & -1.1319 & -1.3303 & 0.4934 \\
MTOS (w/o decay) & -1.4924 & -1.7209 & 0.5558 \\
MTOS (w/o topic-choose) & -1.9936 & -2.1753 & 0.1632 \\
MTOS (w/o all) & -2.1553 & -2.4189 & 0.1324 \\
\bottomrule
\end{tabular}
\label{tab:model_comparison}
\end{table}

\section{Conclusion and Future Work}
This study proposes MTOS to investigate the echo chamber effect during the opinion evolution of individuals in multi-topic environments. By comparing MTOS with the classical single-topic model (SSF), the experiments analyze both continuous evolution metrics, such as the Neighbor Correlation Index and the Echo Chamber Index, and overall comparison metrics, including Polarization, Global Disagreement, and Mean NCI. The results indicate that topic correlation significantly influences group opinion dynamics and the formation of echo chambers: strongly positively correlated topics reinforce consensus and exacerbate echo chamber effects, while weakly positive correlations have a milder impact; negatively correlated topics increase opinion diversity and mitigate echo chambers; and uncorrelated topics disperse attention, substantially enhancing opinion diversity and effectively suppressing echo chamber effects. These findings suggest that topic correlation is a crucial factor modulating echo chamber phenomena in multi-topic settings. Future work will focus on expanding the number of topics and the range of value orientations, optimizing LLMs’ training, and integrating multi-model strategies. This aims to improve the neutrality and diversity of agents’ cognitive expressions. Overall, the MTOS framework offers an extensible experimental platform for future research in public opinion and social simulation.

\bibliographystyle{unsrt}  
\bibliography{references}  


\end{document}